\documentclass[11pt]{article}
\usepackage{acl2016}
\usepackage{times}
\usepackage{latexsym}
\usepackage{mathtools}
\usepackage{comment}
\usepackage{dingbat}

\aclfinalcopy 

\usepackage{algorithm2e}
\usepackage{url}
\usepackage{amsmath}
\usepackage{amsthm}
\usepackage{amssymb}
\usepackage{graphicx}
\usepackage{xspace}
\usepackage{tabularx}
\usepackage{multicol}
\usepackage{multirow}
\usepackage{url}
\usepackage{wrapfig}
\usepackage[utf8]{inputenc}
\usepackage[normalem]{ulem}
\usepackage{xcolor}

\newcommand{\bmx}[0]{\begin{bmatrix}}
\newcommand{\emx}[0]{\end{bmatrix}}

\newcommand{\vect}[1]{\mathbf{#1}}

\newcommand{\matr}[1]{\mathbf{#1}}

\newcommand{\vc}[0]{\vect{c}}
\newcommand{\vh}[0]{\vect{h}}

\newcommand{\vz}[0]{\vect{z}}

\newcommand{\vg}[0]{\vect{g}}

\newcommand{\mW}[0]{\matr{W}}

\newcommand{\ola}{\overleftarrow}
\newcommand{\ora}{\overrightarrow}

\graphicspath{{}}

\title{A Character-Level Decoder without Explicit Segmentation\\ 
       for Neural Machine Translation}

\author{Junyoung Chung \\
        Universit\'e de Montr\'eal \\
        {\small \texttt{junyoung.chung@umontreal.ca}}
	  \And
	Kyunghyun Cho \\
    New York University 
	  \And
	Yoshua Bengio \\
    Universit\'e de Montr\'eal \\
    CIFAR Senior Fellow
  }

\date{}

\begin{document}

\maketitle
\vspace*{-7mm}
\begin{abstract}
    The existing machine translation systems, whether phrase-based or neural,
    have relied almost exclusively on word-level modelling with explicit
    segmentation. In this paper, we ask a fundamental question: can neural
    machine translation generate a character sequence without any explicit
    segmentation? To answer this question, we evaluate an
    attention-based encoder--decoder with a subword-level encoder and a
    character-level decoder on four language pairs--En-Cs, En-De, En-Ru and
    En-Fi-- using the parallel corpora from WMT'15. Our experiments show that
    the models with a character-level decoder outperform the ones with a
    subword-level decoder on all of the four language pairs.
    Furthermore, the ensembles of neural models with a character-level
    decoder outperform the state-of-the-art non-neural machine translation
    systems on En-Cs, En-De and En-Fi and perform comparably on En-Ru.
\end{abstract}

\vspace*{-3mm}
\section{Introduction}
\label{sec:intro}
\vspace*{-2mm}

The existing machine translation systems have relied almost exclusively on
word-level modelling with explicit segmentation. This is mainly due to the issue
of data sparsity which becomes much more severe, especially for $n$-grams, when a sentence is represented
as a sequence of characters rather than words, as the length of the sequence
grows significantly. In addition to data sparsity, we often have a priori belief
that a word, or its segmented-out lexeme, is a basic unit of meaning, making it
natural to approach translation as mapping from a sequence of source-language words to a
sequence of target-language words.

This has continued with the more recently proposed paradigm of neural machine
translation, although neural networks do not suffer from character-level
modelling and rather suffer from the issues specific to word-level modelling,
such as the increased computational complexity from a very large target
vocabulary~\cite{jean2014using,luong2014addressing}.
Therefore, in this paper,
we address a question of whether {\em neural machine translation can be done
directly on a sequence of characters without any explicit word
segmentation}. 

To answer this question, we focus on representing the target side as a character
sequence.  We evaluate neural machine translation models with a character-level
decoder on four language pairs from WMT'15 to make our evaluation as convincing
as possible. We represent the source side as a sequence of subwords extracted
using byte-pair encoding from \newcite{sennrich2015neural}, and vary the target
side to be either a sequence of subwords or characters. On the target side, we
further design a novel recurrent neural network (RNN), called {\em bi-scale recurrent
network}, that better handles multiple timescales in a sequence, and test it in
addition to a naive, stacked recurrent neural network.

On all of the four language pairs--En-Cs, En-De, En-Ru and En-Fi--, the models with a
character-level decoder outperformed the ones with a subword-level decoder.  We
observed a similar trend with the ensemble of each of these configurations,
outperforming both the previous best neural and non-neural translation systems
on En-Cs, En-De and En-Fi, while achieving a comparable result on En-Ru.  We
find these results to be a strong evidence that neural machine translation can
indeed learn to translate at the character-level and that in fact, it benefits
from doing so.

\section{Neural Machine Translation}
\label{sec:nmt}

Neural machine translation refers to a recently proposed approach to machine
translation~\cite{forcada1997recursive,kalchbrenner2013recurrent,Cho-et-al-EMNLP2014,sutskever2014sequence}. 
This approach aims at building an end-to-end neural network that takes as input a source sentence
$X=(x_1,\dots,x_{T_x})$ and outputs its translation $Y=(y_1,\dots,y_{T_y})$,
where $x_t$ and $y_{t'}$ are respectively source and target symbols.  This
neural network is constructed as a composite of an encoder network and a decoder
network.

The encoder network encodes the input sentence $X$ into its continuous
representation. In this paper, we closely follow the neural translation model
proposed in \newcite{bahdanau2014neural} and use a bidirectional recurrent
neural network, which consists of two recurrent neural
networks. The forward network reads the input sentence in a forward
direction:
\mbox{
$
    \ora{\vz}_t = \ora{\phi}(e_x(x_t), \ora{\vz}_{t-1}),
$
}
where $e_x(x_t)$ is a continuous embedding of the $t$-th input symbol, and $\phi$
is a recurrent activation function. Similarly, the reverse network reads the
sentence in a reverse direction (right to left):
\mbox{
    $
    \ola{\vz}_t = \ola{\phi}(e_x(x_t), \ola{\vz}_{t+1}).
    $
}
At each location in the input sentence, we concatenate the hidden states from
the forward and reverse RNNs to form a context set
\mbox{
$
    C = \left\{ \vz_1, \ldots, \vz_{T_x} \right\},
$
}
where $\vz_t = \left[ \ora{\vz}_t; \ola{\vz}_t \right]$.

Then the decoder computes the conditional distribution over all possible
translations based on this context set. This is done by first rewriting the
conditional probability of a translation:
$
    \log p(Y|X) = \sum_{t'=1}^{T_y} \log p(y_{t'} | y_{<t'},
    X).
$
For each conditional term in the summation, the decoder RNN updates its hidden
state by
\begin{align}
    \label{eq:dec_state}
    \vh_{t'} = \phi(e_y(y_{t'-1}), \vh_{t'-1}, \vc_{t'}),
\end{align}
where $e_y$ is the continuous embedding of a target symbol. $\vc_{t'}$ is a
context vector computed by a soft-alignment mechanism:
\begin{align}
    \label{eq:att_context}
    \vc_{t'} = f_{\mathrm{align}}(e_y(y_{t'-1}), \vh_{t'-1}, C)).
\end{align}

The soft-alignment mechanism $f_{\mathrm{align}}$ weights each vector in the
context set $C$
according to its relevance given
what has been translated. 
The weight of each vector $\vz_t$ is computed by
\begin{align}
    \label{eq:att}
    \alpha_{t, t'} = \frac{1}{Z}e^{f_{\mathrm{score}} (e_y(y_{t'-1}), \vh_{t'-1}, \vz_t)},
\end{align}
where $f_{\mathrm{score}}$ is a parametric function returning an unnormalized
score for $\vz_t$ given $\vh_{t'-1}$ and $y_{t'-1}$. We use a feedforward
network with a single hidden layer in this paper.\footnote{
    {\scriptsize For other possible implementations, see \cite{luong2015effective}.}
} $Z$ is a normalization constant:
\mbox{
$
    Z = \sum_{k=1}^{T_x} e^{f_{\mathrm{score}} (e_y(y_{t'-1}), \vh_{t'-1}, \vz_k)}.
$
}
This procedure can be understood as computing the alignment probability between
the $t'$-th target symbol and $t$-th source symbol.

The hidden state $\vh_{t'}$, together with the previous target symbol $y_{t'-1}$
and the context vector $\vc_{t'}$, is fed into a feedforward neural network to
result in the conditional distribution:
\begin{align}
    \label{eq:output}
    p(y_{t'}\mid y_{<t'}, X) \propto
    e^{f^{y_{t'}}_{\mathrm{out}}(e_y(y_{t'-1}), \vh_{t'}, \vc_{t'})}.
\end{align}
The whole model, consisting of the encoder, decoder and soft-alignment
mechanism, is then tuned end-to-end to minimize the negative log-likelihood using stochastic gradient descent.

\section{Towards Character-Level Translation}
\label{sec:motivation}

\subsection{Motivation}

Let us revisit how the source and target sentences ($X$ and $Y$) are represented
in neural machine translation. For the source side of any given training corpus,
we scan through the whole corpus to build a vocabulary $V_x$ of unique tokens to
which we assign integer indices.  A source sentence $X$ is then built as a
sequence of the indices of such tokens belonging to the sentence, i.e., $X=(x_1,
\ldots, x_{T_x})$, where $x_t \in \left\{ 1, 2, \ldots, |V_x|\right\}$. The
target sentence is similarly transformed into a target sequence of integer
indices. 

Each token, or its index, is then transformed into a so-called one-hot vector of
dimensionality $|V_x|$.  All but one elements of this vector are set to 0. The
only element whose index corresponds to the token's index is set to 1. This
one-hot vector is the one which any neural machine translation model sees. The
embedding function, $e_x$ or $e_y$, is simply the result of applying a linear transformation
(the embedding matrix) to this one-hot vector.

The important property of this approach based on one-hot vectors is that the
neural network is oblivious to the underlying semantics of the tokens. To the
neural network, each and every token in the vocabulary is equal distance away
from every other token. The semantics of those tokens are simply {\em learned}
(into the embeddings) to maximize the translation quality, or the log-likelihood of the model.

This property allows us great freedom in the choice of tokens' unit. Neural
networks have been shown to work well with word tokens~\cite{bengio2001neural,schwenk2007continuous,mikolov2010recurrent}
but also with finer units, such as
subwords~\cite{sennrich2015neural,botha2014compositional,luong2013better} as well as
symbols resulting from compression/encoding~\cite{chitnis2015variable}. Although
there have been a number of previous research reporting the use of neural
networks with characters (see, e.g., \newcite{mikolov2012subword} and \newcite{santos2014learning}),
the dominant approach has been to preprocess the text into a sequence of symbols, each associated
with a sequence of characters,
after which the neural network is presented with those symbols rather than
with characters.

More recently in the context of neural machine translation, two research groups
have proposed to directly use characters. \newcite{kim2015character} proposed to
represent each word not as a single integer index as before, but as a sequence
of characters, and use a convolutional network followed by a highway network~\cite{srivastava2015training} to
extract a continuous representation of the word. This approach, which
effectively replaces the embedding function $e_x$, was adopted by
\newcite{costa2016character} for neural machine translation. Similarly,
\newcite{ling2015character} use a bidirectional recurrent neural network to
replace the embedding functions $e_x$ and $e_y$ to respectively encode a
character sequence to and from the corresponding continuous word representation.
A similar, but slightly different approach was proposed by 
\newcite{lee2015naver}, where they
explicitly mark each character with its relative location in a word (e.g., ``B''eginning
and ``I''ntermediate). 

Despite the fact that these recent approaches work at the level of
characters, it is less satisfying that they all rely on knowing how to segment
characters into words. Although it is generally easy for languages like English,
this is not always the case. This word segmentation procedure can be as
simple as tokenization followed by some punctuation normalization, but also can
be as complicated as morpheme segmentation requiring a separate model to be
trained in advance \cite{creutz2005unsupervised,huang2007chinese}. Furthermore,
these segmentation\footnote{
    From here on, the term {\em segmentation} broadly refers to any
    method that splits a given character sequence into a sequence of subword
    symbols.
}
steps are often tuned or designed separately from the ultimate objective of
translation quality, potentially contributing to a suboptimal quality.

Based on this observation and analysis, in this paper, we ask ourselves and the
readers a question which should have been asked much earlier: {\em Is it
possible to do character-level translation without any explicit segmentation?}

\subsection{Why Word-Level Translation?}

\paragraph{(1) Word as a Basic Unit of Meaning}
A word can be understood in two different senses. In the abstract sense, a word
is a basic unit of meaning (lexeme), and in the other sense, can be understood
as a ``concrete word as used in a sentence.''~\cite{booij2012grammar}. A word in
the former sense turns into that in the latter sense via a process of
morphology, including inflection, compounding and derivation. These three
processes do alter the meaning of the lexeme, but often it stays close to the
original meaning.
Because of this view of words as basic units of meaning (either in the form of
lexemes or derived form) from linguistics, much of previous work in natural
language processing has focused on using words as basic units of which a
sentence is encoded as a sequence. Also, the potential difficulty in finding a
mapping between a word's character sequence and meaning\footnote{
    For instance, ``quit'', ``quite'' and ``quiet'' are one edit-distance away
    from each other but have distinct meanings.
}
has likely contributed to this trend toward word-level modelling.

\paragraph{(2) Data Sparsity}
There is a further technical reason why much of previous research on machine
translation has considered words as a basic unit. This is mainly due to the fact
that major components in the existing translation systems, such as language
models and phrase tables, are a count-based estimator of probabilities. In other
words, a probability of a subsequence of symbols, or pairs of symbols, is
estimated by counting the number of its occurrences in a training corpus. This
approach severely suffers from the issue of data sparsity, which is due to a
large state space which grows exponentially w.r.t. the length of subsequences
while growing only linearly w.r.t. the corpus size.
This poses a great challenge to character-level modelling, as any subsequence
will be on average 4--5 times longer when characters, instead of words, are
used. Indeed, \newcite{vilar2007can} reported worse performance when the
character sequence was directly used by a phrase-based machine translation
system. More recently, \newcite{neubig2013substring} proposed a method
to improve character-level translation with phrase-based translation systems,
however, with only a limited success. 

\paragraph{(3) Vanishing Gradient}
Specifically to neural machine translation, a major reason behind the wide
adoption of word-level modelling is due to the difficulty in modelling long-term
dependencies with recurrent neural
networks~\cite{bengio1994learning,hochreiter1998vanishing}. As the lengths of
the sentences on both sides grow 
when they are represented in characters, it is easy to believe that there will
be more long-term dependencies that must be captured by the recurrent neural
network for successful translation. 

\subsection{Why Character-Level Translation?}

\paragraph{Why {\it not} Word-Level Translation?}
The most pressing issue with word-level processing is that we do not have a
perfect word segmentation algorithm for any one language. A perfect segmentation
algorithm needs to be able to segment any given sentence into a sequence of
lexemes and morphemes. This problem is however a difficult problem on its own
and often requires decades of research (see, e.g., \newcite{creutz2005unsupervised}
for Finnish and other morphologically rich languages and \newcite{huang2007chinese}
for Chinese). Therefore, many opt to using either a rule-based tokenization
approach or a suboptimal, but still available, learning based
segmentation algorithm. 

The outcome of this naive, sub-optimal segmentation is that the vocabulary is
often filled with many similar words that share a lexeme but have different
morphology. For instance, if we apply a simple tokenization script to an English
corpus, ``run'', ``runs'', ``ran'' and ``running'' are all separate entries in
the vocabulary, while they clearly share the same lexeme ``run''. This prevents
any machine translation system, in particular neural machine translation, from
modelling these morphological variants efficiently. 

More specifically in the case of neural machine translation, each of these
morphological variants--``run'', ``runs'', ``ran'' and ``running''-- will be
assigned a $d$-dimensional word vector, leading to four independent vectors,
while it is clear that if we can segment those variants into a lexeme and other
morphemes, we can model them more efficiently. For instance, we can have a
$d$-dimensional vector for the lexeme ``run'' and much smaller vectors for ``s''
and``ing''.  Each of those variants will be then a composite of the lexeme
vector (shared across these variants) and morpheme vectors
(shared across words sharing the same suffix, for example)~\cite{botha2014compositional}. This
makes use of distributed representation, which generally yields better generalization, but seems
to require an optimal segmentation, which is unfortunately almost never available.

In addition to inefficiency in modelling, there are two additional negative
consequences from using (unsegmented) words. First, the translation system
cannot generalize well to novel words, which are often mapped to a token reserved for an unknown word.
This effectively ignores any meaning or structure of the word to be
incorporated when translating. Second, even when a lexeme is common and
frequently observed in the training corpus, its morphological variant may not be.
This implies that the model sees this specific, rare morphological variant much
less and will not be able to translate it well.  However, if this rare
morphological variant shares a large part of its spelling with other more common
words, it is desirable for a machine translation system to exploit those common
words when translating those rare variants.

\paragraph{Why Character-Level Translation?}
All of these issues can be addressed to certain extent by directly modelling
characters. Although the issue of data sparsity arises in
character-level translation, it is elegantly addressed by using a parametric
approach based on recurrent neural networks instead of a non-parametric
count-based approach. Furthermore, in recent years, we have learned how to build
and train a recurrent neural network that can well capture long-term
dependencies by using more sophisticated activation functions, such as 
long short-term memory (LSTM) units~\cite{hochreiter1997long} and gated recurrent units~\cite{Cho-et-al-EMNLP2014}.

\newcite{kim2015character} and \newcite{ling2015finding} recently showed that by
having a neural network that converts a character sequence into a word vector,
we avoid the issues from having many morphological variants appearing as
separate entities in a vocabulary.  This is made possible by sharing the
character-to-word neural network across all the unique tokens. A similar
approach was applied to machine translation by \newcite{ling2015character}.

These recent approaches, however, still rely on the availability of a good, if
not optimal, segmentation algorithm. \newcite{ling2015character} indeed states
that ``[m]uch of the prior information regarding morphology, cognates and rare
word translation among others, should be incorporated''.

It however becomes unnecessary to consider these prior information, if we use a
neural network, be it recurrent, convolution or their combination, directly on
the unsegmented character sequence. The possibility of using a sequence of
unsegmented characters has been studied over many years in the field of deep
learning. For instance, \newcite{mikolov2012subword} and
\newcite{sutskever2011generating} trained a recurrent neural network language
model (RNN-LM) on character sequences. The latter showed that it is possible to
generate sensible text sequences by simply sampling a character at a time from
this model. More recently, \newcite{zhang2015character} and
\newcite{xiao2016efficient} successfully applied a convolutional net and a
convolutional-recurrent net respectively to character-level document
classification without any explicit segmentation.
\newcite{gillick2015multilingual} further showed that it is possible to
train a recurrent neural network on unicode bytes, instead of characters or words,
 to perform part-of-speech tagging and named entity recognition. 

These previous works suggest the possibility of applying neural networks for the task of machine translation,
which is often considered a substantially more difficult problem compared to
document classification and language modelling.

\subsection{Challenges and Questions}

There are two overlapping sets of challenges for the source and target sides. On
the source side, it is unclear how to build a neural network that learns a
highly nonlinear mapping from a spelling to the meaning of a sentence.  

On the target side, there are two challenges. The first challenge is the same
one from the source side, as the decoder neural network needs to summarize what
has been translated. 
In addition to this, the character-level modelling on the target side is more
challenging, as the decoder network must be able to generate a long, coherent
sequence of characters. This is a great challenge, as the size of the state
space grows exponentially w.r.t. the
number of symbols, and in the case of characters, it is often 300-1000 symbols
long.

All these challenges should first be framed as questions; whether the current
recurrent neural networks, which are already widely used in neural machine
translation, are able to address these challenges as they are. In this paper, we
aim at answering these {\em questions} empirically and focus on the challenges
on the target side (as the target side shows both of the challenges).

\section{Character-Level Translation}
\label{sec:char_trans}

In this paper, we try to answer the questions posed earlier by testing two
different types of recurrent neural networks on the target side (decoder).

First, we test an existing recurrent neural network with gated recurrent units
(GRUs). We call this decoder a {\em base} decoder.

Second, we build a novel two-layer recurrent neural network, inspired by the
gated-feedback network from \newcite{chung2015gated}, called a {\em bi-scale}
recurrent neural network. 
We design this network to facilitate capturing two timescales, motivated by the
fact that characters and words may work at two separate timescales.

We choose to test these two alternatives for the following purposes. Experiments
with the base decoder will clearly answer whether the existing neural network is
enough to handle character-level decoding, which has not been properly answered
in the context of machine translation. The alternative, the bi-scale decoder, is
tested in order to see whether it is possible to design a better decoder, if the
answer to the first question is positive.

\begin{figure}
    \begin{minipage}{1.\columnwidth}
        \begin{minipage}{0.46\columnwidth}
            \centering
            \includegraphics[width=1.\columnwidth]{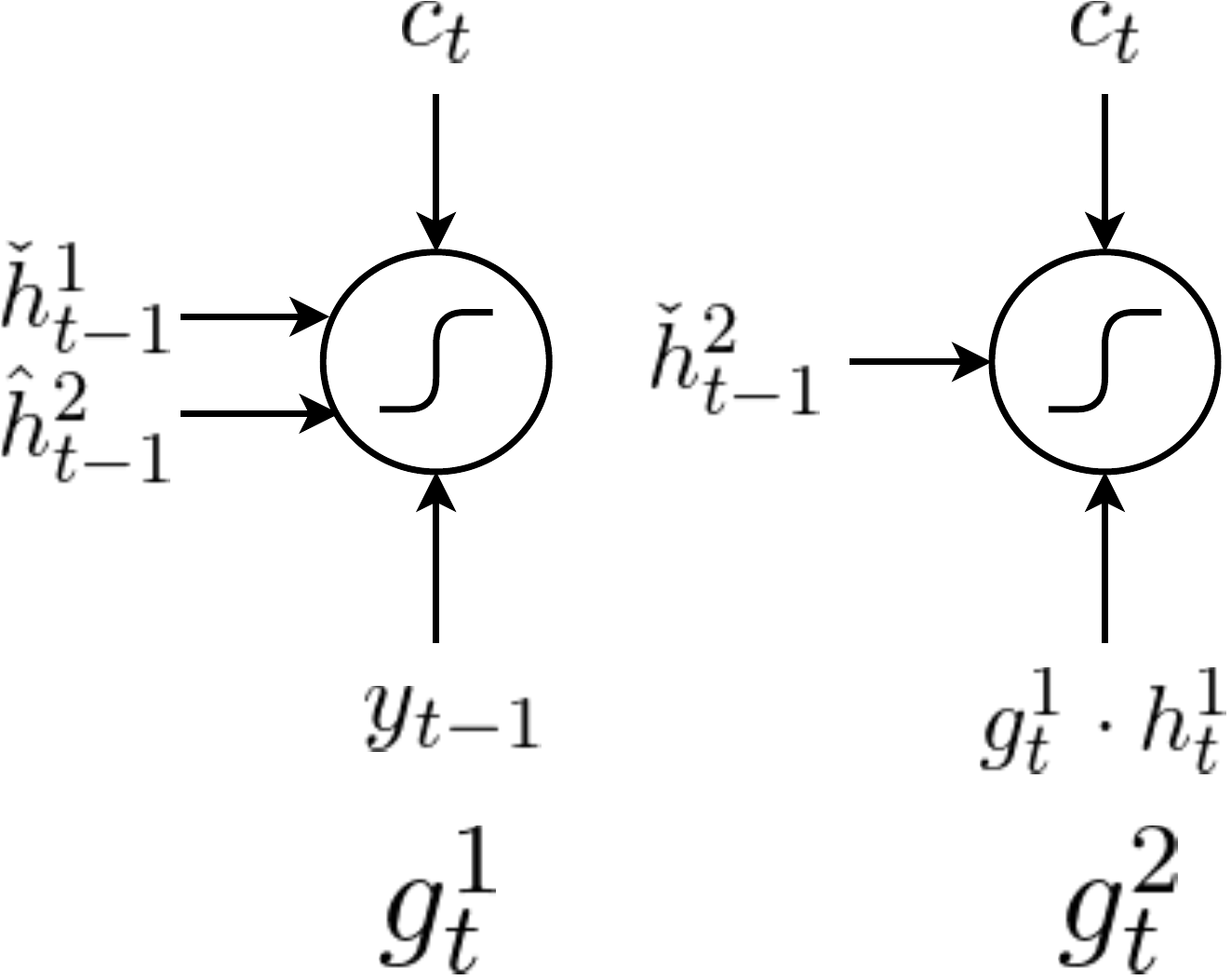}
        \end{minipage}
        \hfill
        \begin{minipage}{0.46\columnwidth}
            \centering
            \includegraphics[width=1.\columnwidth]{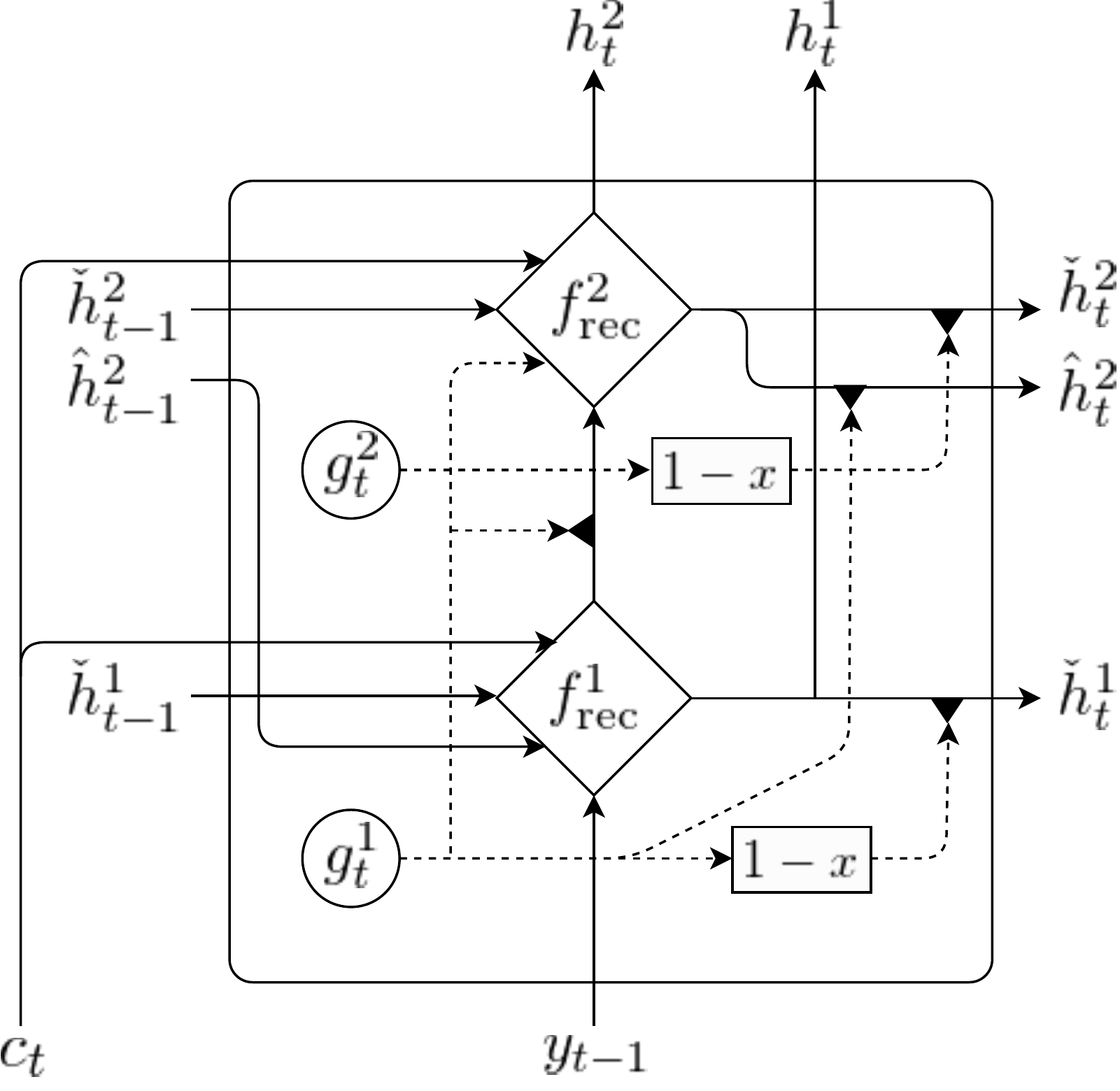}
        \end{minipage}
    \end{minipage}
    \begin{minipage}{1.\columnwidth}
        \begin{minipage}{0.48\columnwidth}
            \centering
            (a) Gating units
        \end{minipage}
        \begin{minipage}{0.48\columnwidth}
            \centering
            (b) One-step processing
        \end{minipage}
    \end{minipage}

    \caption{Bi-scale recurrent neural network}
    \label{fig:CHDec}

\end{figure}

\subsection{Bi-Scale Recurrent Neural Network}

In this proposed bi-scale recurrent neural network, there are two sets of hidden
units, $\vh^1$ and $\vh^2$. They contain the same number of units, i.e.,
$\text{dim}(\vh^1) = \text{dim}(\vh^2)$. The first set $\vh^1$ models a
fast-changing timescale (thereby, a {\em faster layer}), and $\vh^2$ a slower
timescale (thereby, a {\em slower layer}). For each hidden unit, there is an
associated gating unit, to which we refer by $\vg^1$ and $\vg^2$. For the
description below, we use $y_{t'-1}$ and $\vc_{t'}$ for the previous target symbol
and the context vector (see Eq.~\eqref{eq:att_context}), respectively.

Let us start with the faster layer. The faster layer outputs two sets of
activations, a normal output $\vh^1_{t'}$ and its gated version
$\check{\vh}^1_{t'}$. The activation of the faster layer is computed by
\begin{align*}
    \vh^1_{t'} =& \tanh\left( 
    \mW^{h^1} \left[ e_y(y_{t'-1}); \check{\vh}^1_{t'-1};
    \hat{\vh}^2_{t'-1}; \vc_{t'}\right] 
    \right),
\end{align*}
where $\check{\vh}^1_{t'-1}$ and $\hat{\vh}^2_{t'-1}$ are the gated activations
of the faster and slower layers respectively.  These gated activations are
computed by
\begin{align*}
    \check{\vh}^1_{t'} = (1 - \vg^1_{t'}) \odot \vh^1_{t'},~~\hat{\vh}^2_{t'} = \vg^1_{t'} \odot \vh^2_{t'}.
\end{align*}

In other words, the faster layer's activation is based on the adaptive
combination of the faster and slower layers' activations from the previous time
step. Whenever the faster layer determines that it needs to reset, i.e.,
$\vg^1_{t'-1} \approx 1$, the next activation will be determined based more on
the slower layer's activation. 

The faster layer's gating unit is computed by
\begin{align*}
    \vg^1_{t'} = \sigma\left( 
        \mW^{g^1} \left[ e_y(y_{t'-1}); 
        \check{\vh}^1_{t'-1}; \hat{\vh}^2_{t'-1}; \vc_{t'}\right] 
    \right),
\end{align*}
where $\sigma$ is a sigmoid function.

The slower layer also outputs two sets of activations, a normal output
$\vh^2_{t'}$ and its gated version $\check{\vh}^2_{t'}$. These activations are 
computed as follows: 
\begin{align*}
    &\vh^2_{t'} = 
    (1 - \vg^1_{t'}) \odot \vh^2_{t'-1} +
    \vg^1_{t'} \odot \tilde{\vh}^2_{t'}, \\
    &\check{\vh}^2_{t'} = (1 - \vg^2_{t'}) \odot \vh^2_{t'},
\end{align*}
where $\tilde{\vh}^2_{t'}$ is a candidate activation. The slower layer's gating
unit $\vg^2_{t'}$ is computed by
\begin{align*}
    \vg^2_{t'} =& \sigma\left(\mW^{g^2}
    \left[(\vg^1_{t'} \odot \vh^1_{t'}) ;
    \check{\vh}^2_{t'-1};
\vc_{t'}\right]\right).
\end{align*}

This adaptive leaky integration based on the gating unit from the faster layer
has a consequence that the slower layer updates its activation only when the
faster layer resets. This puts a soft constraint that the faster layer runs at a
faster rate by preventing the slower layer from updating while the faster layer
is processing a current chunk.

The candidate activation is then computed by
\begin{align}
    \label{eq:slow_act}
    \hspace*{-0.1cm}
    \tilde{\vh}^2_{t'} = 
    \tanh\left(
    \mW^{h^2}\left[(\vg^1_{t'}\odot\vh^1_{t'}); \check{\vh}^2_{t'-1}; \vc_{t'}\right]\right).
\end{align}

$\check{\vh}^2_{t'-1}$ indicates the {\em reset} activation from the previous time step,
similarly to what happened in the faster layer, 
and $\vc_{t'}$ is the input from the context. 

According to $\vg^1_{t'}\odot\vh^1_{t'}$ in Eq.~\eqref{eq:slow_act}, the faster layer
influences the slower layer, only when the faster layer has finished processing
the current chunk and is about to {\em reset} itself ($\vg^1_{t'} \approx 1$).
In other words, the slower layer does not receive any input from the faster
layer, until the faster layer has quickly processed the current chunk, thereby
running at a slower rate than the faster layer does.

At each time step, the final output of the proposed bi-scale recurrent neural
network is the concatenation of the output vectors of the faster and slower
layers, i.e., $\left[ \vh^1; \vh^2 \right]$. This concatenated vector is used to
compute the probability distribution over all the symbols in the vocabulary, as
in Eq.~\eqref{eq:output}.  See Fig.~\ref{fig:CHDec} for graphical illustration.

\begin{figure}[t]
    \begin{minipage}{0.49\columnwidth}
        \centering
        \includegraphics[width=1.\columnwidth]{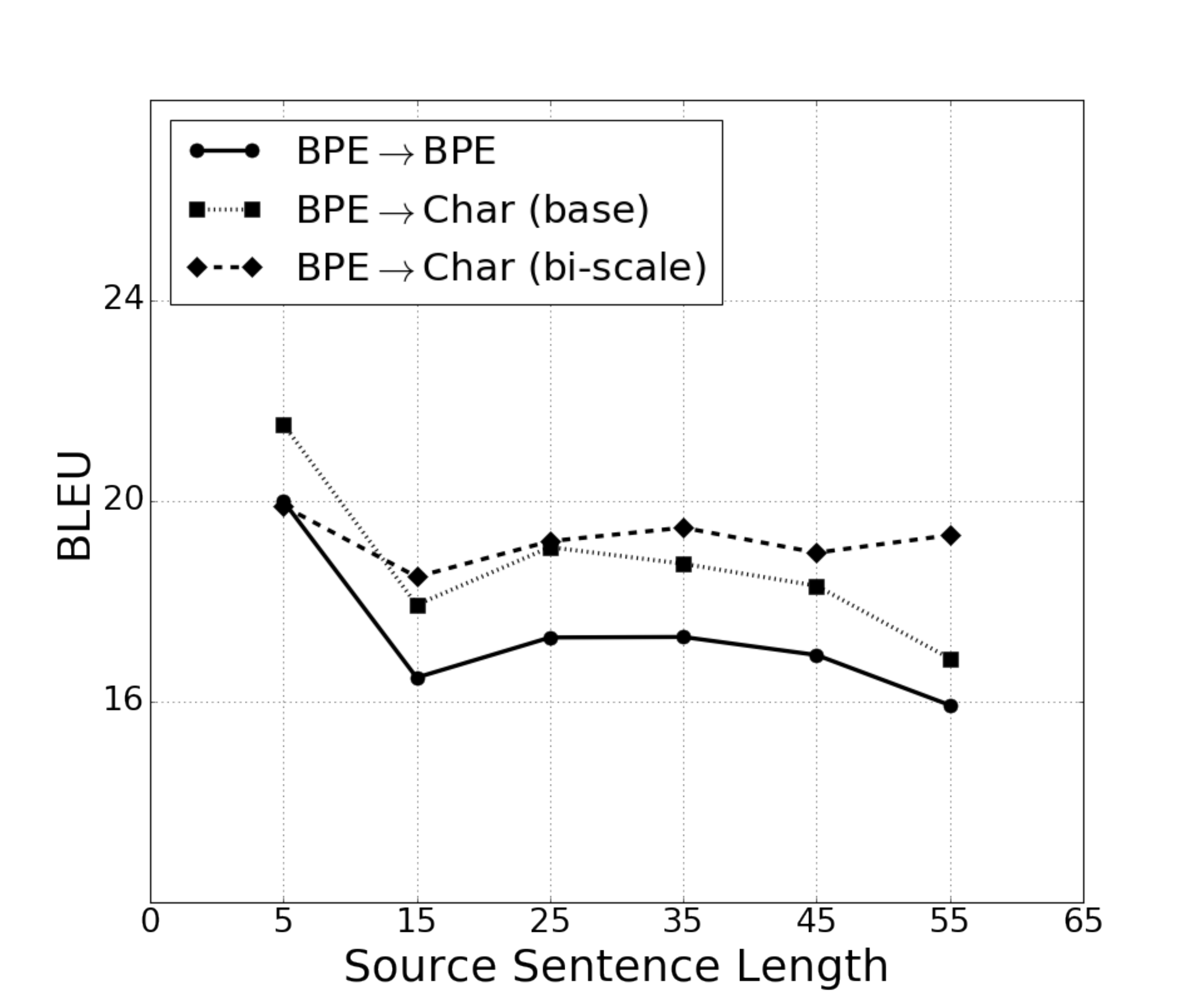}
    \end{minipage}
    \hfill
    \begin{minipage}{0.49\columnwidth}
        \centering
        \includegraphics[width=1.\columnwidth]{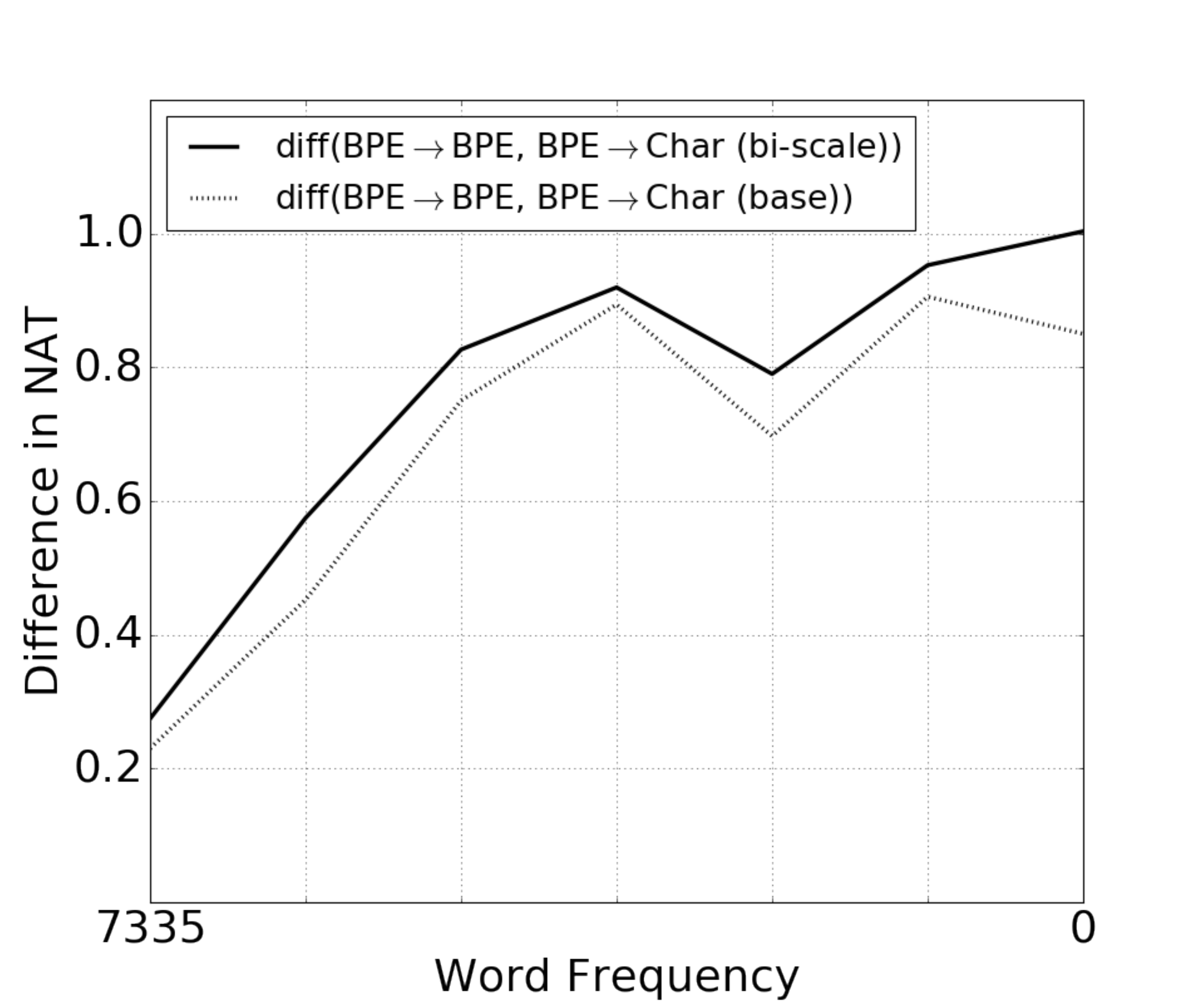}
    \end{minipage}

    \caption{(left) The BLEU scores on En-Cs w.r.t. the length of source sentences. 
        (right) The difference of word negative log-probabilities
        between the subword-level decoder and either of the character-level base
        or bi-scale decoder. 
    }
    \label{fig:qual1}

\end{figure}

\begin{table*}[ht]
    \small
    \vfill
    \centering
    \begin{tabular}{c | c | c | c | c | c | c | c || c | c || c | c || c |  c }
        & 
        & \multirow{2}{*}{\rotatebox[origin=c]{90}{Src}} & &
        \multirow{2}{*}{\rotatebox[origin=c]{55}{Depth}}       &
        \multicolumn{2}{c|}{Attention} & 
        \multirow{2}{*}{\rotatebox[origin=c]{55}{Model}} & 
        \multicolumn{2}{c||}{Development} &
        \multicolumn{2}{c||}{$\mathrm{Test}_1$} & 
        \multicolumn{2}{c}{$\mathrm{Test}_2$} \\
        \cline{6-7}\cline{9-14}
        & & & Trgt           &  & $\vh^1$ & $\vh^2$ & &
        Single & Ens & 
        Single & Ens & 
        Single & Ens \\
        \hline
        \hline
        \multirow{9}{*}{\rotatebox[origin=c]{90}{En-De}} & (a) &
        \multirow{9}{*}{\rotatebox[origin=c]{90}{BPE}} & \multirow{2}{*}{BPE} & 1 & \checkmark & & \multirow{2}{*}{Base}  
        & 20.78 & --
        & 19.98 & --
        & 21.72 & -- \\
        &          (b) &                 &                  & 2     & \checkmark & \checkmark    &            
        & $21.26_{20.62}^{21.45}$ & 23.49 
        & $20.47_{19.30}^{20.88}$ & 23.10
        & $22.02_{21.35}^{22.21}$ & 24.83 \\
        \cline{4-14}
        &          (c) &                  & \multirow{6}{*}{Char} & 2     &            & \checkmark    & \multirow{2}{*}{Base} 
        & $21.57_{20.88}^{21.88}$ & 23.14 
        & $\mathbf{21.33}_{19.82}^{21.56}$ & \underline{23.11}
        & $\mathbf{23.45}_{21.72}^{23.91}$ & 25.24 \\
        &          (d) &                  &                  & 2     & \checkmark & \checkmark    &            
        & 20.31 &   --
        & 19.70 & --     
        & 21.30 & --      \\
        \cline{5-14}
        &          (e) &                  &                  & 2     &
        & \checkmark    & \multirow{3}{*}{Bi-S} 
        & $21.29_{21.13}^{21.43}$ & 23.05
        & $21.25_{20.62}^{21.47}$ & 23.04
        & $23.06_{22.85}^{23.47}$ &  \underline{25.44}  \\
        &          (f) &                  &                  & 2     & \checkmark & \checkmark    &            
        & 20.78 & -- 
        & 20.19 & --      
        & 22.26 & --      \\
        &          (g) &                  &                  & 2     & \checkmark &               &            
        & 20.08 & -- 
        & 19.39 & --      
        & 20.94 & --      \\
        \cline{5-14}
        \cline{2-14}
        & \multicolumn{7}{c||}{State-of-the-art Non-Neural Approach$^\ast$}
        & \multicolumn{2}{c||}{--} & \multicolumn{2}{c||}{20.60$^{(1)}$} &
        \multicolumn{2}{c}{24.00$^{(2)}$}       \\
        \hline
        \hline
        \multirow{4}{*}{\rotatebox[origin=c]{90}{En-Cs}} & (h) &
         \multirow{3}{*}{\rotatebox[origin=c]{90}{BPE}} & BPE & 2 & \checkmark & \checkmark & Base  
        & $16.12_{15.96}^{16.96}$ & 19.21 
        & $17.16_{16.38}^{17.68}$ & 20.79
        & $14.63_{14.26}^{15.09}$ & 17.61 \\
        \cline{4-14}
        &          (i) &                  & \multirow{2}{*}{Char} & 2&            & \checkmark    & Base       
        & $17.68_{17.39}^{17.78}$ & 19.52
        & $19.25_{18.89}^{19.55}$ & 21.95
        & $\mathbf{16.98}_{16.81}^{17.17}$ & 18.92 \\
        \cline{5-14}
        &          (j) &                  &                  & 2     &
        & \checkmark    & Bi-S   
        & $17.62_{17.43}^{17.93}$ & 19.83
        & $\mathbf{19.27}_{19.15}^{19.53}$ & \underline{22.15}
        & $16.86_{16.68}^{17.10}$ & \underline{18.93} \\
        \cline{2-14}
        & \multicolumn{7}{c||}{State-of-the-art Non-Neural Approach$^\ast$}
        & \multicolumn{2}{c||}{--} & \multicolumn{2}{c||}{21.00$^{(3)}$} &
        \multicolumn{2}{c}{18.20$^{(4)}$}       \\
        \hline
        \hline
        \multirow{4}{*}{\rotatebox[origin=c]{90}{En-Ru}} & (k) &
         \multirow{3}{*}{\rotatebox[origin=c]{90}{BPE}} & BPE & 2 & \checkmark & \checkmark & Base  
        & $18.56_{18.26}^{18.70}$ & 21.17
        & $25.30_{24.95}^{25.40}$ & 29.26 
        & $19.72_{19.02}^{20.29}$ & 22.96 \\
        \cline{4-14}
        &          (l)       &            & \multirow{2}{*}{Char} & 2&            & \checkmark    & Base       
        & $18.56_{18.39}^{18.87}$ & 20.53
        & $\mathbf{26.00}_{25.04}^{26.07}$ & \underline{29.37}
        & $\mathbf{21.10}_{20.14}^{21.24}$ & 23.51 \\
        \cline{5-14}
        &          (m)     &              &                  & 2     &
        & \checkmark    & Bi-S   
        & $18.30_{17.88}^{18.54}$ & 20.53
        & $25.59_{24.57}^{25.76}$ & 29.26
        & $20.73_{19.97}^{21.02}$ & \underline{23.75} \\
        \cline{2-14}
        & \multicolumn{7}{c||}{State-of-the-art Non-Neural Approach$^\ast$}
        & \multicolumn{2}{c||}{--} & \multicolumn{2}{c||}{28.70$^{(5)}$} &
        \multicolumn{2}{c}{24.30$^{(6)}$}       \\
        \hline
        \hline
        \multirow{4}{*}{\rotatebox[origin=c]{90}{En-Fi}} & (n) &
         \multirow{3}{*}{\rotatebox[origin=c]{90}{BPE}} & BPE & 2 & \checkmark & \checkmark & Base  
        & $9.61_{9.24}^{10.02}$ & 11.92 
        & -- &  --
        & $8.97_{8.88}^{9.17}$ & 11.73 \\
        \cline{4-14}
        &          (o)   &                & \multirow{2}{*}{Char} & 2&            & \checkmark    & Base       
        & $11.19_{11.09}^{11.55}$ & 13.72
        & --    & --
        & $\mathbf{10.93}_{10.11}^{11.56}$ & \underline{13.48} \\
        \cline{5-14}
        &          (p) &                  &                  & 2     &
        & \checkmark    & Bi-S   
        & $10.73_{10.40}^{11.04}$ & 13.39
        & --    & --
        & $10.24_{9.71}^{10.63}$ & 13.32 \\
        \cline{2-14}
        & \multicolumn{7}{c||}{State-of-the-art Non-Neural Approach$^\ast$}
        & \multicolumn{2}{c||}{--} & \multicolumn{2}{c||}{--} &
        \multicolumn{2}{c}{12.70$^{(7)}$}       \\
        \hline
    \end{tabular}
    \caption{BLEU scores of the subword-level, character-level base and
    character-level bi-scale decoders for both single models and ensembles. The best
    scores among the single models per language pair are bold-faced, and those
    among the ensembles are underlined.
    When available, we report the
    median value, and the minimum and maximum values as a subscript and a
    superscript, respectively. 
    {\small ($\ast$) 
        \url{http://matrix.statmt.org/} as of 11 March 2016 (constrained only).
             (1) \newcite{freitag2014eu}. 
             (2, 6) \newcite{williams2015edinburgh}.
             (3, 5) \newcite{durrani2014edinburgh}.
             (4) \newcite{haddow2015edinburgh}.
             (7) \newcite{rubino2015abu}.}
     }
    \label{tab:single_models}

\end{table*}

\section{Experiment Settings}
\label{sec:experiments}

For evaluation, we represent a source sentence as a sequence of subword symbols
extracted by byte-pair encoding (BPE, \newcite{sennrich2015neural}) and a target
sentence either as a sequence of BPE-based symbols or as a sequence of
characters. 

\paragraph{Corpora and Preprocessing}
We use all available parallel corpora for four language pairs from WMT'15:
En-Cs, En-De, En-Ru and En-Fi. They consist of 12.1M, 4.5M, 2.3M
and 2M sentence pairs, respectively. We tokenize each corpus using a tokenization
script included in Moses.\footnote{
    Although tokenization is not necessary for character-level modelling, we
    tokenize the all target side corpora to make comparison against word-level
    modelling easier. 
} 
We only use the sentence pairs, when the source side is up to 50 subword symbols
long and the target side is either up to 100 subword symbols or 500 characters.
We do not use any monolingual corpus.

For all the pairs other than En-Fi, we use newstest-2013 as a development
set, and newstest-2014 (Test$_1$) and newstest-2015 (Test$_2$) as test sets.
For En-Fi, we use newsdev-2015 and newstest-2015 as development and test sets,
respectively.

\paragraph{Models and Training}
We test three models settings: (1) BPE$\to$BPE, (2) BPE$\to$Char (base) and (3)
BPE$\to$Char (bi-scale). The latter two differ by the type of recurrent neural
network we use. 
We use GRUs for the encoder in all the settings. We used GRUs for the decoders in the
first two settings, (1) and (2), while the proposed bi-scale recurrent network was used in
the last setting, (3). The encoder has $512$ hidden units for each direction (forward
and reverse), and the decoder has $1024$ hidden units per layer. 

We train each model using stochastic gradient descent with 
Adam~\cite{kingma2014adam}. Each update is computed using a minibatch of 128
sentence pairs. The norm of the gradient is clipped with a threshold
$1$~\cite{pascanu2013construct}. 

\paragraph{Decoding and Evaluation}
We use beamsearch to approximately find the most likely translation given a
source sentence. The beam widths are $5$ and $15$ respectively for the
subword-level and character-level decoders. They were chosen based on the
translation quality on the development set. The translations are
evaluated using BLEU.\footnote{
    We used the multi-bleu.perl script from Moses.
}

\paragraph{Multilayer Decoder and Soft-Alignment Mechanism}

When the decoder is a multilayer recurrent neural network (including a stacked
network as well as the proposed bi-scale network), the decoder outputs multiple
hidden vectors--$\left\{\vh^1, \ldots, \vh^L\right\}$ for $L$ layers, at a time.
This allows an extra degree of freedom in the soft-alignment mechanism
($f_{\mathrm{score}}$ in Eq.~\eqref{eq:att}). We evaluate using alternatives,
including (1) using only $\vh^L$ (slower layer) and
(2) using all of them
(concatenated).

\paragraph{Ensembles}

We also evaluate an ensemble of neural machine translation models and compare
its performance against the state-of-the-art phrase-based translation systems on
all four language pairs. We decode from an ensemble by taking the
average of the output probabilities at each step.

\begin{figure*}[ht]
    \begin{minipage}{0.6\textwidth}
        \centering
        \includegraphics[height=3.3cm,clip=True,trim=110 0 110 0]{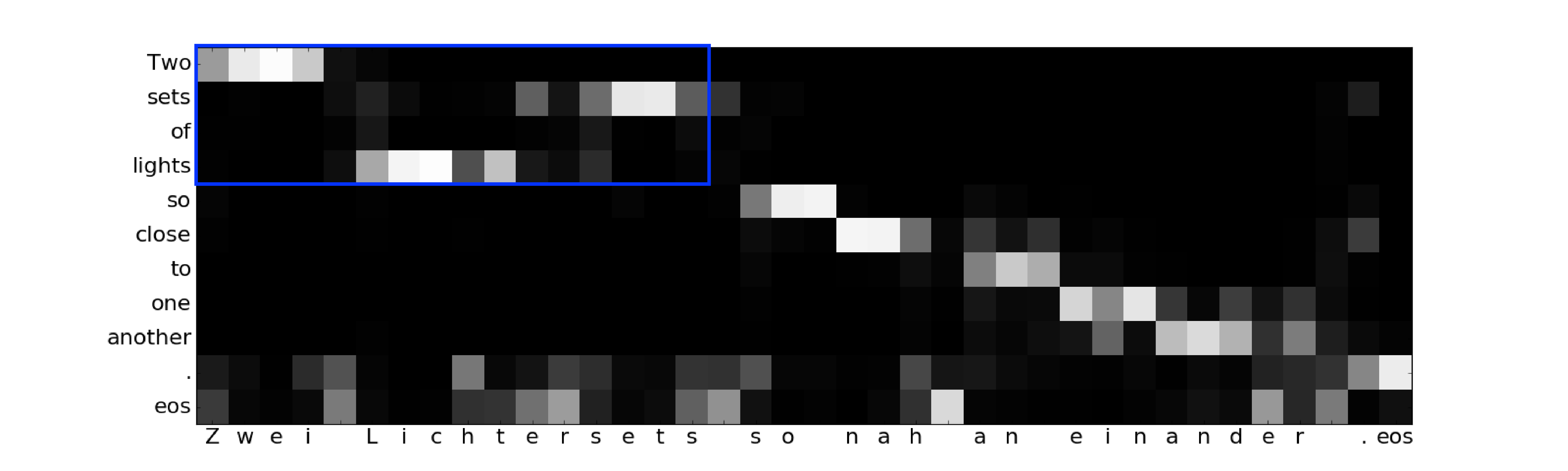}
    \end{minipage}
    \begin{minipage}{0.3\textwidth}
        \includegraphics[height=3.3cm,clip=True,trim=40 0 40 0]{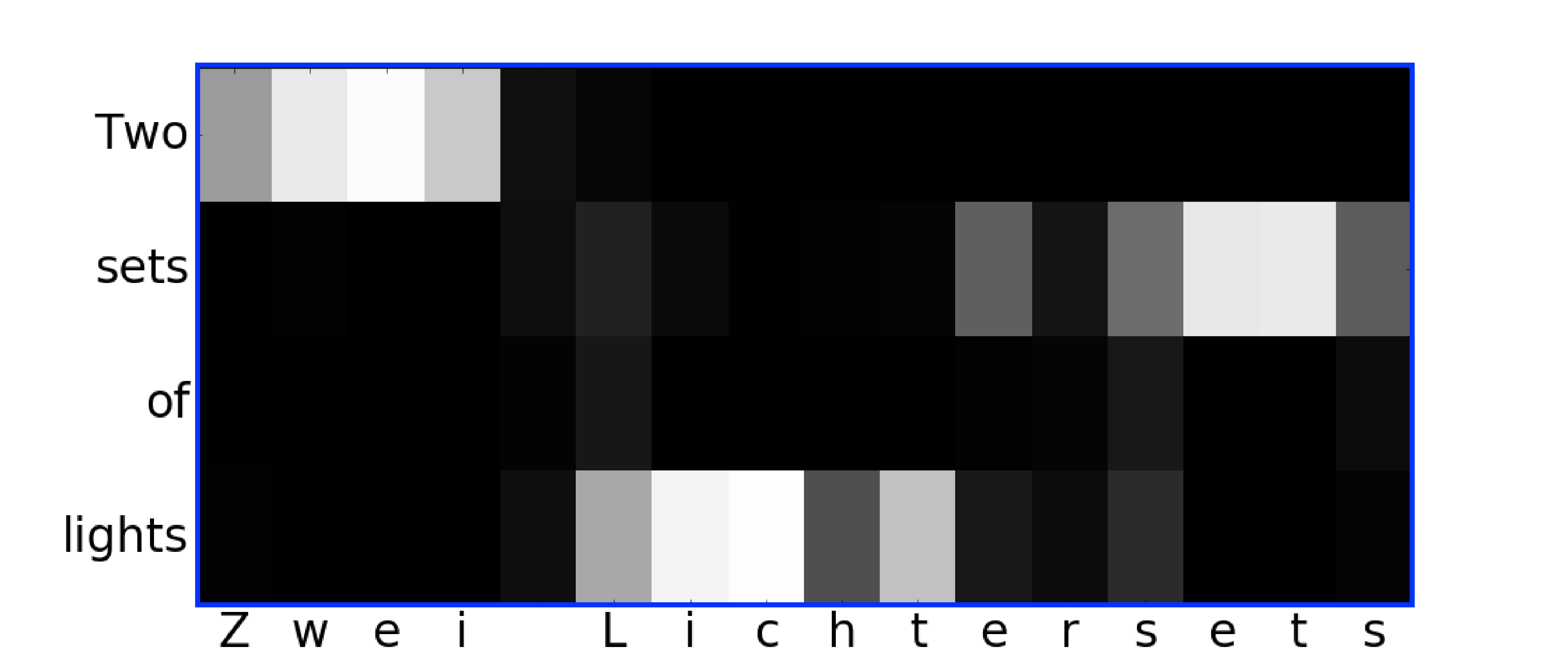}
    \end{minipage}
    \hfill
    \caption{Alignment matrix of a test example from En-De
             using the BPE$\rightarrow$Char (bi-scale) model.}
    \label{fig:align}

\end{figure*}

\section{Quantitative Analysis}

\paragraph{Slower Layer for Alignment}
On En-De, we test which layer of the decoder should be used for computing
soft-alignments. In the case of subword-level decoder, we observed no
difference between choosing any of the two layers of the decoder against using
the concatenation of all the layers (Table~\ref{tab:single_models}~(a--b)) On
the other hand, with the character-level decoder, we noticed an improvement when
only the slower layer ($\vh^2$) was used for the soft-alignment mechanism
(Table~\ref{tab:single_models}~(c--g)).  This suggests that the soft-alignment
mechanism benefits by aligning a larger chunk in the target with a subword unit
in the source, and we use only the slower layer for all the other language
pairs.

\paragraph{Single Models}
In Table~\ref{tab:single_models}, we present a comprehensive report of the
translation qualities of (1) subword-level decoder, (2) character-level base
decoder and (3) character-level bi-scale decoder, for all the language pairs. We
see that the both types of character-level decoder outperform the subword-level
decoder for En-Cs and En-Fi quite significantly. On En-De, the
character-level base decoder outperforms both the subword-level
decoder and the character-level bi-scale decoder, validating the
effectiveness of the character-level modelling.
On En-Ru, among the single models, the character-level decoders
outperform the subword-level decoder, but in general, we observe that all the three
alternatives work comparable to each other. 

These results clearly suggest that
{\em it is indeed possible to do character-level translation without
explicit segmentation.} In fact, what we observed is that character-level
translation often surpasses the translation quality of
word-level translation. Of course, we note once again that our experiment is
restricted to using an unsegmented character sequence at the decoder only, and a
further exploration toward replacing the source sentence with an unsegmented
character sequence is needed.

\paragraph{Ensembles}
Each ensemble was built using eight independent models.
The first observation we make is that in all the language
pairs, neural machine translation performs comparably to, or often better than,
the state-of-the-art non-neural translation system. Furthermore, the
character-level decoders outperform the subword-level decoder in all the cases.

\section{Qualitative Analysis}

\paragraph{(1) Can the character-level decoder generate a long, coherent
sentence?}
The translation in characters is dramatically longer than that in words, likely
making it more difficult for a recurrent neural network to generate a coherent
sentence in characters. This belief turned out to be false.  As shown in
Fig.~\ref{fig:qual1}~(left), there is no significant difference between the
subword-level and character-level decoders, even though the lengths of the
generated translations are generally 5--10 times longer in characters.

\paragraph{(2) Does the character-level decoder help with rare words?}
One advantage of character-level modelling is that it can model the composition
of any character sequence, thereby better modelling rare morphological variants.
We empirically confirm this by observing the growing gap in the average negative
log-probability of words between the subword-level and character-level decoders
as the frequency of the words decreases. This is shown in
Fig.~\ref{fig:qual1}~(right) and explains one potential cause behind the success of
character-level decoding in our experiments (we define $\mathrm{diff}(x, y) = x - y$).

\paragraph{(3) Can the character-level decoder soft-align between a source
word and a target character?}
In Fig.~\ref{fig:align}~(left), we show an example soft-alignment of a source
sentence, ``Two sets of light so close to one another''. It is clear that the
character-level translation model well captured the alignment between the source
subwords and target characters. We observe that the character-level decoder
correctly aligns to ``lights'' and ``sets of'' when generating a German compound
word ``Lichtersets'' (see Fig.~\ref{fig:align}~(right) for the zoomed-in version).
This type of behaviour happens similarly between ``one another'' and
``einander''. Of course, this does not mean that there exists an alignment
between a source word and a target character.  Rather, this suggests that the
internal state of the character-level decoder, the base or bi-scale, well
captures the meaningful chunk of characters, allowing the model to map it to a larger chunk
(subword) in the source.

\paragraph{(4) How fast is the decoding speed of the character-level decoder?}
We evaluate the decoding speed of subword-level base, character-level base and 
character-level bi-scale decoders on newstest-2013 corpus (En-De) with a single Titan X GPU.
The subword-level base decoder generates 31.9 words per second, and the character-level base decoder
and character-level bi-scale decoder generate 27.5 words per second and 25.6 words per second, respectively.
Note that this is evaluated in an online setting, performing consecutive translation, where only one sentence is translated
at a time. Translating in a batch setting could differ from these results.

\section{Conclusion}
\label{sec:conclusion}

In this paper, we addressed a fundamental question on whether a recently
proposed neural machine translation system can directly handle translation at
the level of characters without any word segmentation. We focused on the target
side, in which a decoder was asked to generate one character at a time, while
soft-aligning between a target character and a source subword.
Our extensive experiments,
on four language pairs--En-Cs, En-De, En-Ru and En-Fi-- strongly
suggest that it is indeed possible for neural machine translation to translate
at the level of characters, and that it actually benefits from doing so.

Our result has one limitation that we used subword symbols in the source side.
However, this has allowed us a more fine-grained analysis, but in the future, a
setting where the source side is also represented as a character sequence must
be investigated.

\section*{Acknowledgments}
\label{sec:ack}
The authors would like to thank the developers of Theano~\cite{team2016theano}.
We acknowledge the support of the following agencies for research funding and
computing support: NSERC, Calcul Qu\'{e}bec, Compute Canada,
the Canada Research Chairs, CIFAR and Samsung. KC thanks the support by
Facebook, Google (Google Faculty Award 2016) and NVIDIA (GPU Center of Excellence 2015-2016).
JC thanks Orhan Firat for his constructive feedbacks.

\bibliography{junyoung}
\bibliographystyle{acl2016}
\end{document}